\documentclass[lettersize,journal]{IEEEtran}
\usepackage{amsmath,amsfonts}
\usepackage{algorithm}
\usepackage{array}

\usepackage[font=normalsize]{subfig}
\usepackage{textcomp}
\usepackage{stfloats}
\usepackage{url}
\usepackage{verbatim}
\usepackage{graphicx}
\usepackage{cite}

\usepackage{booktabs}

\usepackage{caption}
\usepackage{multirow}
\usepackage{etoolbox}
\makeatletter
\patchcmd{\@makecaption}
  {\scshape}
  {}
  {}
  {}
\makeatletter
\patchcmd{\@makecaption}
  {\\}
  {.\ }
  {}
  {}
\makeatother

\usepackage{tipa}

\usepackage{caption}

\usepackage{subfig}

\usepackage{algorithm}
\usepackage{algorithmicx}
\usepackage{algpseudocode}
\usepackage{verbatim}
\usepackage{graphicx}
\usepackage{multirow}

\graphicspath{ {images/} }

\AtBeginDocument{%
  }

\begin{document}

\title{REWAFL: Residual Energy and Wireless Aware Participant Selection for Efficient Federated Learning over Mobile Devices}


\author{Yixuan Li,~\IEEEmembership{Student Member,~IEEE}, Xiaoqi Qin,~\IEEEmembership{Member,~IEEE}, Jiaxiang Geng,~\IEEEmembership{Student Member,~IEEE}, Rui Chen, Yanzhao Hou, Yanmin Gong,~\IEEEmembership{Member,~IEEE}, Miao Pan,~\IEEEmembership{Senior Member,~IEEE}, and Ping Zhang,~\IEEEmembership{Fellow,~IEEE}
\thanks{Y. Li, X. Qin, J. Geng, Y. Hou and P. Zhang are with the State Key Laboratory of Networking and Switching Technology,  Beijing University of Posts and Telecommunications, Beijing, 100876, China, (e-mail: lyixuan@bupt.edu.cn; xiaoqiqin@bupt.edu.cn; lelegjx@bupt.edu.cn; houyanzhao@bupt.edu.cn; pzhang@bupt.edu.cn). P. Zhang is also with the Department of Broadband Communication, Peng Cheng Laboratory, Shenzhen 518066, Guangdong, China.

R. chen and M. Pan are with the Department of Electrical and Computer Engineering, University of Houston, Houston, TX 77204 USA (e-mail: rchen19@uh.edu; mpan2@uh.edu).

Y. Gong is with the Department of Electrical and Computer Engineering, University of Texas at San Antonio, San Antonio, TX 78249
USA (e-mail: yanmin.gong@utsa.edu).
}

}


\maketitle

\begin{abstract}
Participant selection (PS) helps to accelerate federated learning (FL) convergence, which is essential for the practical deployment of FL over mobile devices. While most existing PS approaches focus on improving training accuracy and efficiency rather than residual energy of mobile devices, which fundamentally determines whether the selected devices can participate. Meanwhile, the impacts of mobile devices' heterogeneous wireless transmission rates on PS and FL training efficiency are largely ignored. Moreover, PS causes the staleness issue. Prior research exploits isolated functions to force long-neglected devices to participate, which is decoupled from original PS designs.

In this paper, we propose a \underline{r}esidual \underline{e}nergy and \underline{w}ireless \underline{a}ware PS design for efficient \underline{FL} training over mobile devices (REWAFL). REWAFL introduces a novel PS utility function that jointly considers global FL training utilities and local energy utility, which integrates energy consumption and residual battery energy of candidate mobile devices. Under the proposed PS utility function framework, REWAFL further presents a residual energy and wireless aware local computing policy. Besides, REWAFL buries the staleness solution into its utility function and local computing policy. The experimental results show that REWAFL is effective in improving training accuracy and efficiency, while avoiding ``flat battery'' of mobile devices.
\end{abstract}


\section{Introduction}
\IEEEPARstart{N}{owadays}, federated learning (FL) has been migrating from cable-powered datacenters to battery-powered mobile devices~\cite{Pan21infocom}. With ever advancing hardware development, mobile devices (e.g., Google Pixel 4a, Galaxy Note20, iPad Pro, MacBook laptop, UAVs, etc.)
have increasing on-device computing capability ready for local training. Besides, mobile devices crowdsense the raw training data and provide the foundation for FL beyond datacenters \cite{Zhu20MCOM, Qin22JSAC}. Following FL principle of training deep neural networks (DNNs) locally without exposing raw data, FL over mobile devices fosters numerous promising applications in various domains. For example, Google and Apple have deployed FL for computer vision and natural language processing tasks across mobile devices \cite{Cho2020ClientSI, 2017LearningWP,  geyer2019differentially, oort34}; NVIDIA exploits FL to create medical imaging AI \cite{Fan2020Sol}; many other applications~\cite{li2020federated, nguyen2021federated, Ye20} are emerging in e-Healthcare, autonomous driving, hazard detection in smart home, smart surveillance and monitoring in industrial Internet of Things (IoT), etc. Most of those applications require to train FL
models across a crowd of clients, while mobile devices may not all be simultaneously available or suitable for FL training. Therefore, how to appropriately select FL participants is the key to their successful deployment in practice.

The biggest challenges for FL participant selection (PS) stem from the heterogeneity of mobile devices. To improve the performance of  FL over mobile devices, research efforts in the literature tried to address the data heterogeneity and system heterogeneity issues, and developed FL PS approaches considering \emph{statistical utility} or/and \emph{system utility} as defined in~\cite{OortOSDI2021}. For example, to address data heterogeneity/improve statistical utility, the importance-aware scheduling policies are proposed to select mobile devices with significant training improvement. Various importance metrics for mobile devices are defined based on gradient divergence \cite{Ren20Scheduling}, local loss values or gradient norm~\cite{Tianyi18NIPS,Katharopoulos2018NotAS,Salehi18Sampling}, and the differences between the local and global models~\cite{Wu22noniid}.
To address system heterogeneity/reduce FL training latency, the mobile devices with the best instantaneous channel conditions at each round are selected to minimize the communication latency~\cite{Zhu20Analog, Amiri2021Scheduling}. To improve the efficiency of model aggregation, the central server evaluates the computing capability at model devices and select participants with the ``fine'' models \cite{Ye20}.
Pioneeringly, Oort~\cite{OortOSDI2021} proposed a utility function that jointly considers statistical and system heterogeneity to guide PS for efficient FL training. While most existing PS approaches focus on improving learning accuracy and reducing training latency from the perspective of global FL system,
there are very limited discussions about concerns from participating mobile devices' side,
i.e., the energy consumption of FL training and the residual-battery heterogeneity among candidate mobile devices. It is not trivial since if the selected device cannot participate in FL training, it will backfire FL performance in terms of accuracy and convergence latency.
Note that only reducing system-level energy consumption~\cite{kim2021autofl, RuiEEFL, lyx22TVT} may not be enough, since the mobile device with very low residual battery energy is not feasible for selection, even though its energy consumption for training is small.

Besides, state-of-the-art (SOTA) designs largely ignore the impacts of wireless transmission heterogeneity on local computing policy so as to reduce latency/energy consumption, and its consequent influence on PS in FL training. It is obvious that given a FL task, fast/slow transmissions will decrease/increase the communication latency/energy consumption of uploading local model updates from a mobile device to FL server. However, it is not clear when a mobile device's wireless transmission rate is high/low, how to adjust its own local computing policy over rounds. Additionally, it is not clear how these adjustments affect local computing latency/energy consumption considering device's residual energy, and further PS and FL performance.

Moreover, since not every mobile device participates in every round of FL training, frequently selecting a fixed subset of devices to participate may lead to parameter staleness and biased training, which eventually degrades training performance. Most prior PS designs enforce those high-staleness mobile devices to participate and update their model parameters by a separated function~\cite{OortOSDI2021}, which is decoupled from their original PS designs.
In addition, existing staleness solutions ignore the selected participating devices' energy consumption and its residual battery energy.

Motivated by the challenges above, in this paper, we propose a \underline{r}esidual \underline{e}nergy and \underline{w}ireless \underline{a}ware PS design for efficient FL training over mobile devices (REWAFL). Briefly, REWAFL introduces a novel PS utility function that jointly considers global FL training utilities ( statistical utility and global latency utility) and local energy utility, which is aware of energy consumption and residual battery energy of candidate participating mobile devices. Under the proposed PS utility function framework, REWAFL further presents a residual energy and wireless aware local computing policy, which includes a wireless aware local stochastic gradient descent (SGD) computing strategy, and an energy utility aware stopping criterion for increasing local training iterations. Besides, REWAFL buries the staleness solution into its utility function and local computing development policy without incurring any extra or external mechanism.

To demonstrate its effectiveness, we implemented and evaluated a prototype of REWAFL system consisting of FL server (NVIDIA RTX 3090), mobile devices (Android smartphones, tablet computers, and laptops) with heterogeneous battery energy and different wireless transmission techniques (Wi-Fi 5 and 5G) and rates for model updates between them. We conducted extensive experiments with various FL learning tasks, training model and datasets. Our experimental results show that the proposed REWAFL can reduce device dropout ratio, reach the target accuracy with less overall latency and energy consumption, and effectively solve the staleness issue in a self-contained manner compared with SOTA designs.

\section{PRELIMINARIES AND MOTIVATION}
\subsection{Federated Learning over Mobile Devices}

We consider a wireless federated learning system consisting of an edge server (e.g., base station or gNodeB) as the FL aggregator and a set of $\mathcal{S}$ mobile devices as FL clients. We assume that the computing resources (e.g., GPU and CPU frequencies), wireless transmission rates (i.e., $s$), data distribution and residual battery energy (i.e., $E$) among mobile devices are heterogeneous. The edge server and mobile devices jointly execute FL training as follows. The edge server first broadcasts an initialized global model to the participating mobile devices. After receiving the global model, each selected mobile device $i$ utilizes local computing resources to train its local models by performing $H(i)$ local iterations based on its local data of size $|B_i|$. When the local training is complete, mobile devices upload model parameters to the edge server
using available wireless accessing technologies, i.e., Wi-Fi 5, 5G, etc. After that, the edge server aggregates the local model to update the global model and selects the participants for the next round based on the defined PS utility function.
The procedure above repeats until FL converges, where PS plays an essential role for FL's learning performance and efficiency.

\subsection{
Revisiting SOTA PS Designs
}
\label{sec:Limitations}
Some pioneering FL PS designs in the literature considered not only FL learning performance, but also FL training efficiency in terms of global system latency or local participating mobile devices' energy consumption.
For example, Oort in~\cite{OortOSDI2021} proposed the PS design to associate the statistical utility (i.e., contributions to FL learning performance) and system efficiency (i.e., system latency, which is the sum of local computing latency and communication latency). In Oort, mobile device $i$'s utility function for PS is formulated as follows.

\begin{equation}
\begin{split}
Util(i) = |B_i|\sqrt{\frac{1}{|B_i|}\sum_{\substack{k\in {B}_i}}Loss(k)^2}
\times (\frac{T}{t(i)})^{\mathbb{I}(T<t(i)) \times \alpha}  ,
\end{split}
\label{Eq:oortUtil}
\end{equation}
where $B_i$ denotes the local training samples of mobile device $i$, $Loss(k)$ denotes the training loss of the $k$-th sample, $t(i)$ denotes the system latency of mobile device $i$, $T$ is the developer-preferred duration of each round and $\alpha$ is the penalty factor. Besides, $\mathbb{I}(x)$ is an indicator function, where $\mathbb{I}(x)=1$ if $x$ is true, and 0 otherwise.

Another example is AutoFL in~\cite{kim2021autofl}, which proposed a PS design to improve learning performance while reducing total energy consumption. Briefly, AutoFL introduced a reward function to associate learning accuracy and the energy consumption of mobile devices, and exploited reinforcement learning to select participants with high contributions to FL training and low energy consumption.

Although Oort/AutoFL exhibits good learning performance and global latency/local energy consumption efficiency, SOTA PS approaches ignored the battery energy constraints and various transmission conditions confronted by candidate mobile devices in FL training, i.e., (i) the battery-powered mobile devices may have heterogeneous levels of residual energy, and (ii) mobile devices have to face heterogeneous wireless transmission environments. Such ignorance may downgrade the training performance/efficiency of FL over mobile devices, or even make the selected mobile devices unable to participate in FL training.

\subsection{
Unawareness of Heterogeneous Residual Battery Energy
}
\label{sec:Battery}
Mobile devices are powered by the battery, which has a limited energy capacity. If the FL PS design is unaware of mobile devices' residual energy heterogeneity, it may quickly deplete the energy of frequently selected mobile devices. Since a mobile device has to reserve certain amount of energy for its mandatory/regular operations (e.g., display, voice calls, data services, location services, etc.), such designs may drain out the energy of frequently selected devices and disable them from participation afterwards in FL training. That will have detrimental effects on FL learning performance and efficiency, especially when there is a heterogeneous data distribution among candidate devices.

In order to understand the impact of residual  battery energy heterogeneity on training efficiency, we conducted empirical experiments to validate our projection. Here, we follow SOTA designs, i.e., Oort in~\cite{OortOSDI2021} and AutoFL in~\cite{kim2021autofl}, to select participants in each training round. The training performances are acquired on mobile devices, i.e., Android smartphones, tablet and laptop. We set NVIDIA RTX 3090 as the FL server.
We consider a hybrid communication scenario with 5G and Wi-Fi 5.
More details of the experimental setup can be found in Section V.

\begin{table}[]
\caption{The Dropout Ratio of SOTA PS Designs
   }
   \centering
\begin{tabular}{|l|c|c|c|}
\hline
\multirow{2}{*}{Local Model} & \multirow{2}{*}{\begin{tabular}[c]{@{}c@{}}Target\\ Acc. (\%) \end{tabular}} & \multicolumn{2}{|c|}{Dropout Ratio (\%)}
\\ \cline{3-4}
                               &                                                                            & \begin{tabular}[c]{@{}c@{}}Oort\end{tabular} & \begin{tabular}[c]{@{}c@{}}AutoFL\end{tabular} \\ \hline
CNN@HAR                        & 89.3                                                                   & 85.0                                                   & 55.0                                                   \\ \hline
CNN@CIFAR10                    & 72.2                                                                   & 46.0                                                   & 25.0                                                   \\ \hline
LSTM@Shakespeare               & 50.3                                                                   & 66.0                                                   & 53.0                                                   \\ \hline
\end{tabular}
\label{Tabel:motivation}
\end{table}

\begin{figure}[tbp]
	\centering
		\centering
		\includegraphics[width=0.36\textwidth]{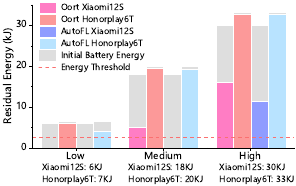}
   \caption{Energy consumption and residual energy of devices with different initial energy (CNN@MNIST).}
	\label{Figure:motivation}
\end{figure}

First,  we define the dropout ratio as the ratio of the number of mobile devices that cannot participate in FL training due to depleted battery energy and the total number of candidate mobile devices.
From the results in Table~\ref{Tabel:motivation},
we observe that both Oort and AutoFL have very high dropout ratio to reach the target accuracy under different learning tasks. The potential reason is that the SOTA PS designs are not aware of candidate mobile devices' residual energy, and thus may drain out the batteries of frequently selected devices.
Then, for CNN@MNIST, we take the high-end mobile devices (Xiaomi 12S smartphone with Adreno 730 GPU) with high transmission rates (i.e., 79.60Mbps for 5G) and the low-end devices (Honor Play 6T smartphone with Mali-G57 MC2 GPU) with low transmission rates (i.e., 0.64Mbps for 5G) as examples to present how much energy is left for mobile devices with different initial battery energy levels after FL training.
From the results in Fig.~\ref{Figure:motivation}, we find that high-end mobile devices use a lot or even use up its battery energy, while low-end devices don't use much. It indicates that high-end devices are much more frequently selected for training than low-end ones, in order to reduce global latency~\cite{OortOSDI2021} or energy consumption~\cite{kim2021autofl}. Again, such designs may deplete the energy of frequently selected devices with low initial battery energy at early rounds, and exclude them from participation at later training rounds, which may backfire FL learning performance and latency/energy efficiency.

\subsection{
Unawareness of Heterogeneous Wireless Transmission Rates
}
\label{sec:hetertx}
Mobile devices confront various wireless transmission environments, and thus have heterogeneous transmission rates. During each FL training round,
mobile device $i$' wireless transmission rate $s(i)$ explicitly affects its energy consumption,
training latency and residual energy, all of which are important metrics for FL PS. As we know, the energy consumption/latency of mobile device $i$ per FL training round consists of two parts: on-device local computing energy consumption/latency, and communication energy consumption/latency for uploading local model updates. Obviously, for a given FL task, the high/low transmission rate $s(i)$ reduces/increases both the energy consumption and latency of mobile device $i$ for uploading the model updates to FL server.

However, it is not clear what the corresponding local computing policy should be, when the wireless transmission rate is high/low. Questions to answer include: (i) How to adjust local computing to be aware of the candidate mobile device's transmission rate?
(ii) How to adjust local computing to be aware of the candidate device's residual energy? (iii) How does the local computing policy affect PS and FL performance? In order to achieve good learning performance and training efficiency, a fine-grained local computing policy under the FL PS framework is in need.

\subsection{
Limitations of Oort's Staleness Solution
}
\label{sec:Stalenesslimitation}
Frequently selecting a fixed subset of mobile devices to participate may result in the staleness issue in FL training.
Oort in~\cite{OortOSDI2021} provided a solution by introducing an extra ``temporal uncertainty'' mechanism, i.e., if mobile device $i$ has not been selected for a long time, its statistical utility will be forced to increase by adding an independent component characterized by the round counter and mobile device $i$'s last involved round. Such a staleness solution is isolated from the PS utility function. Besides, it ignores their participating devices' energy consumption and its residual battery energy.
It would be desirable if (i) the staleness solution can be aware of participating devices' energy consumption, wireless transmission rate, residual energy levels, etc., and (ii) the staleness issue can inherently be addressed by the PS design in FL training.


\section{ALGORITHM DESIGN}
\label{sec:DESIGN}

\begin{figure*} \centering
  {\includegraphics[width=1.0\linewidth]{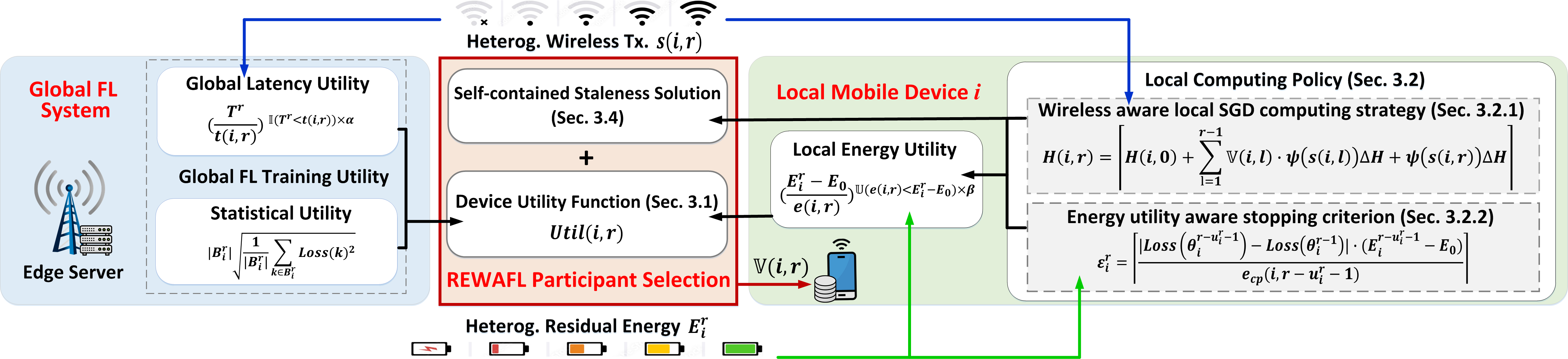}}
  \vspace{-4mm}
 \caption{The design sketch of residual energy and wireless aware federated learning (REWAFL).} \label{Figure:overview}
 \vspace{-4.3mm}
\end{figure*}

Aiming to address the concerns of both global FL system utility and local mobile devices' energy constraints,
we propose a REWAFL scheme that includes: (i) a \underline{r}esidual \underline{e}nergy \underline{a}ware (REA) PS utility function design, (ii) a fine-grained REWA local computing policy under the proposed PS utility function framework, and (iii) a self-contained REWA staleness solution. 
The sketch of the REWAFL design is shown in Fig.~\ref{Figure:overview}.

\subsection{REAFL: REA PS Utility Function Design
}
\label{sec:UtilityFunction}
As illustrated in Sec.~\ref{sec:Battery}, for FL over mobile devices,
it may not be adequate for PS utility function to just consider the learning performance while reducing energy consumption or/and latency of FL, which may drain out the battery of frequently selected mobile devices. The PS utility function design has to be aware of the residual energy of candidate mobile devices.

Our intuitive idea behind the residual energy aware PS utility function design is to trade-off and jointly improve statistical utility, global system/latency efficiency and energy efficiency in FL training, while satisfying the residual energy constraints of mobile devices.
Rooting from Oort's utility function~\cite{OortOSDI2021} in Eqn.~(\ref{Eq:oortUtil}), we propose the residual-energy aware PS utility function as follows\footnote{Note that the global latency utility in Eqn.~(\ref{Eq:babeUtil}) is the same as the global system utility defined in~\cite{OortOSDI2021}. We rename them to (i) explicitly distinguish global FL training utility from local mobile device's energy utility, and (ii) jointly consider statistical, global latency and device's energy utilities.}.

\begin{small}
\begin{equation}
\begin{split}
Util(i,r) &= \overbrace{\underbrace{|B_i^r|\sqrt{\frac{1}{|B_i^r|}\sum_{\substack{k\in {B}_i^r}}Loss(k)^2}}_{\textbf{Statistical utility}}  \times \underbrace{\big(\frac{T^r}{t(i,r)}\big)^{\mathbb{I}(T^r<t(i,r)) \times \alpha}}_{\textbf{Global latency utility}}}^{\textbf{
Global FL training utility
}} \\
&\times \underbrace{\big(\frac{E_i^r-E_0}{e(i,r)}\big)^{\mathbb{U}(e(i,r)<E_i^r-E_0)\times \beta}}_{\textbf{Local mobile device's energy utility}},
\end{split}
\label{Eq:babeUtil}
\end{equation}
\end{small}
where $|B_i^r|$ is the number of training data samples of mobile device $i$ at the $r$-th round,
$t(i,r)$ is the training latency of mobile device $i$, $T^r$ is the developer-preferred duration at $r$th round \cite{OortOSDI2021}, $e(i,r)$ is the energy consumption of mobile device $i$ at the $r$-th round, $E_i^r$ is residual battery energy of mobile device $i$ at the $r$-th round, $E_0$ is the threshold energy reserved for its mandatory/regular operations,
and thus $E_i^r-E_0$ is the available energy to consume.
Moreover, $\mathbb{I}(x)$ is an indicator function, where $\mathbb{I}(x)=1$, if $x$ is true, and 0 otherwise.
$\mathbb{U}(x)$ is also an indicator function that takes value 1 if $x$ is true, and $\infty$ otherwise.
Here, $\alpha$ and $\beta$ are scaling coefficients to balance/associate different metrics/utilities in terms of the model accuracy, latency and energy efficiency.

Different from the existing PS utility function designs (e.g., Oort~\cite{OortOSDI2021} or AutoFL~\cite{kim2021autofl}), the proposed utility function considers not only the global FL training utility including statistical utility and global latency utility (similar to those in Oort~\cite{OortOSDI2021}), but also the local mobile device's energy utility, which is aware of heterogeneous residual battery energy of candidate mobile devices.
Here is a simple interpretation of the energy utility.
During the $r$-th round, if the energy consumption of mobile device $i$ is less than its residual available battery energy, i.e., $e(i,r) < E_i^r-E_0$, the energy utility of mobile device $i$ is proportional to the residual available battery energy and inversely proportional to the energy consumption. That makes mobile devices with more residual energy and less energy consumption easier to be selected for participation.
Otherwise, if the energy consumption of mobile device $i$ is greater than or equal to its residual available energy, i.e., $e(i,r) \geq E_i^r-E_0$, the value of the energy utility sharply drops to zero and mobile device $i$ will not be able to join model training in the $r$-th round.
Therefore, the energy efficiency of the proposed energy utility component has two-fold meanings, i.e., low energy consumption and residual energy awareness.

Following the proposed utility function in Eqn.~(\ref{Eq:babeAdapH}), at the beginning of the $r$-th FL training round, the edge server collects the reported information (i.e., $Loss$, $t(i,r)$, and device $i$'s energy utility)\footnote{As for the privacy preservation of $Loss$ and $t(i,r)$, we follow the same analysis in Oort~\cite{OortOSDI2021}. For the privacy of $e(i,r)$ and $E_i^r$, (i) $e(i,r)$ and $E_i^r$ are not that related to the training data, which is sensitive and worth privacy protection, (ii) the client can self-calculate its local energy utility and report it without disclosing $e(i,r)$ and $E_i^r$, and (iii) the client can report the estimated energy utility, or even dishonestly report its energy utility, while the benefits of such dishonest reporting and the consequent games between the client and the server or among clients are out of the scope of this paper.}
from mobile devices and calculate their utilities to select the participants for this round.
Similar to that in Oort~\cite{OortOSDI2021}, we exploit mobile devices' most recent aggregate training losses to estimate their current ones.
Besides, mobile device $i$ needs to estimate its $t(i,r)$ and local energy utility, where $t(i,r)$/$e(i,r)$ consists of local computing latency/energy consumption and communication latency/energy consumption.
Given a FL learning task, assuming device $i$'s transmission power is fixed, the size of device $i$'s local model update transmitted to FL server does not change over global training rounds. Thus, the communication latency/energy consumption of mobile devices can be calculated given a known transmission rate $s(i,r)$.
Moreover, since device $i$'s computing capability is fixed, the computing latency/energy consumption can be approximately estimated by the product\footnote{We neglect the non-linear impacts~\cite{RuiEEFL} brought by advanced hardware latency/energy saving techniques (e.g., DVFS) for local on-device training.} of the latency/energy consumption of each iteration and the number of local iterations $H(i,r)$ at the $r$-th round.
Thus, the estimation of $t(i,r)$ and device $i$'s energy utility relies on $H(i,r)$.
Under the proposed PS utility function framework, in the following subsection, we further illustrate the wireless aware local computing policy (i.e., how to adjust $H(i,r)$) to further improve the energy/latency efficiency for FL training over mobile devices.

\subsection{
REWA Local Computing Policy
}
\label{sec:AdaHPolicy}
The proposed REWA local computing policy for PS includes two parts: (i) a wireless aware local stochastic gradient descent (SGD) computing strategy, and (ii) an energy utility aware stopping criterion for $H(i)$.

\subsubsection{A wireless aware local SGD computing strategy}
Let us start with existing adaptive local SGD computing strategy, AdaH in~\cite{Haddadpour19AdaH}, which gradually increases the number of local iterations over rounds,
i.e., $H(i,r)= \left \lceil H(0) + \sum\limits_{l = 1}^{r} \psi\Delta H \right \rceil$.
Here, $H(i,r)$ is the number of local training iterations for mobile device $i$ at the $r$-th round, $H(0)$ is the initial value of the number of local iterations, $\psi$ is the growth-rate coefficient, and $\Delta H$ is a non-negative increment unit. The rationale behind it is as follows. At the beginning of the training process, the learning performance is less sensitive to the number of local iterations. Even if the local model is updated with fewer local iterations in the early rounds, learning performance is barely affected.
However, as FL training proceeds,
it is more sensitive to local model quality, and thus requires more local training iterations for better learning accuracy.
By using such an AdaH strategy, the statistical utility in Eqn.~(\ref{Eq:babeUtil}) can be effectively improved.

To tailor the AdaH strategy~\cite{Haddadpour19AdaH} for FL PS, we introduce a binary variable $\mathbb{V}(i,r)$ to represent the selection status of mobile device $i$ at the $r$-th round. If mobile device $i$ is selected, $\mathbb{V}(i,r)$ = $1$, and 0 otherwise. Considering participant selection, we can update the local AdaH in \cite{Haddadpour19AdaH} to $H(i,r)=\left \lceil H(i,0) + \sum\limits_{l = 1}^{r-1}  \mathbb{V}(i,l) \cdot \psi \Delta H + \psi \Delta H \right \rceil$.
Thus, the relationship between the local iteration number of the $r$-th round and that of the last participation round can be characterized as follows.
If the mobile device $i$ is selected to participate in training at the current $r$-th round, $H(i,r)$ will increase based on $H(i,r-u_i^r-1)$ at device $i$'s latest participating round, i.e., $H(i,r) = \left \lceil H(i,r-u_i^r-1) + \psi \Delta H \right \rceil$, where $u_i^r$ is the number of non-participating rounds between the current $r$-th round and mobile device $i$'s latest participating round.
Otherwise, $H(i,r) = H(i,r-u_i^r-1)$.

Furthermore, to make the local SGD computing strategy aware of wireless communications, we refine it as follows. 
\begin{equation}
\begin{split}
& H(i,r)
= \left \lceil H(i,r-u_i^r-1) + \psi(s(i,r)) \Delta H \right \rceil\\
&= \left \lceil  H(i,0) + \sum\limits_{l = 1}^{r-1}  \mathbb{V}(i,l) \cdot \psi(s(i,l)) \Delta H + \psi(s(i,r)) \Delta H\right \rceil,
\end{split}
\label{Eq:babeAdapH}
\end{equation}
where $s(i,r)$ is the averaged wireless transmission rate from device $i$ to the edge server at the $r$-th round, and $\psi(\cdot)$ is a non-negative function that decreases with the wireless transmission rate.
For the $r$-th round,
if transmission rate $s(i,r)$ is high,
the increment of $H(i,r)$ will be small; otherwise, if $s(i,r)$ is low,
the increment of $H(i,r)$ will be large.

The rationale behind the strategy in Eqn.~(\ref{Eq:babeAdapH}) is threefold. First, since it inherits ``increasing $H$ over rounds'' strategy from AdaH~\cite{Haddadpour19AdaH} and is tailored for PS, it helps to increase the learning accuracy and \emph{statistical utility}. Second, such a wireless aware local SGD computing strategy makes device $i$ with higher transmission rate at the $r$-th round have smaller increase of local iterations, i.e., less computing workload added. Thus, for a given local learning task, it helps to reduce device $i$'s local computing latency in the $r$-th round of FL training. Since the communication latency has already been reduced due to the fast transmission rate, the wireless aware local computing strategy helps to increase the \emph{global latency utility} for mobile device $i$. Third, once $H(i,r)$ is determined, device $i$'s local computing energy consumption can be estimated for the $r$-th round. Similar to computing latency, small increase of local iterations leads to small increase of local computing energy consumption for device $i$ with high transmission rate. Meanwhile, assuming device $i$'s wireless transmission power is fixed, the energy consumption of local gradient updates transmissions will be reduced for device $i$, if its transmission rate is high. Therefore, the wireless aware local computing strategy in Eqn.~(\ref{Eq:babeAdapH}) helps to reduce the energy consumption of device $i$ with fast transmission and may increase its \emph{energy utility}, which is jointly determined by both the energy consumption and the residual energy of device $i$. With such a wireless aware local computing strategy, mobile devices with high transmission rate may have good utility as defined in Eqn.~(\ref{Eq:babeUtil}), so that they may have good chances to be selected for participation.

Although the proposed wireless aware local computing strategy seems to favor the devices with fast transmissions, it actually also creates participation opportunities for the devices with slow transmissions, and addresses the staleness issue in an inherent manner. The details of staleness analysis and solution will be provided in Sec.~\ref{Sec:Staleness}.

\subsubsection{An energy utility aware stopping criterion for $H(i,r)$}\label{sec:stoppingH}
As $H(i,r)$ keeps increasing over rounds, the computing energy consumption of device $i$ per round will increase in parallel. An unstoppable increase of $H(i,r)$ may ultimately drain out mobile device's battery energy. That prompts us to develop a stopping criterion for $H(i,r)$'s increase, which is aware of device $i$'s energy utility (i.e., energy consumption and residual energy).
Our idea for the energy utility aware stopping criterion for increasing $H(i)$ is simple.
When increase $H(i,r)$ fails to notably reduce the value of the loss function but consumes a lot of computing energy instead, the increasing of $H(i,r)$ should be stopped. Besides, if the residual battery energy of mobile device $i$ is low, we should avoid further increase of $H(i,r)$, allowing the mobile device to retain enough energy for its regular operations.
Following this idea, we propose the energy utility aware stopping criterion for increasing $H(i,r)$ as follows.
\begin{equation}
\begin{split}
\varepsilon_i^r = \frac{|Loss({\boldsymbol{\theta}_i^{r-u_i^r-1}})-Loss({\boldsymbol{\theta}^{r-1}})|\cdot (E_i^{r-u_i^r-1}-E_0)}{e_{cp}(i,r-u_i^r-1)} ,
\end{split}
\label{Eq:AdapHstop}
\end{equation}
where ${\boldsymbol{\theta}_i^{r-u_i^r-1}}$ denotes the local model parameters at the last participating round, i.e., the $(r-u_i^r-1)$-th round, ${\boldsymbol{\theta}^{r-1}}$ denotes the global model parameters at the $(r-1)$-th round,
and $e_{cp}(i,r-u_i^r-1)$ is the computing energy consumption of mobile device $i$ at the ($r-u_i^r-1$)-th round.
Here, if $\varepsilon_i^r$ is large,
we will continue increasing $H(i,r)$ following Eqn.~(\ref{Eq:babeAdapH});
otherwise, if $\varepsilon_i^r$ is smaller than a predefined threshold $\varepsilon_{th}^r$, we will stop increasing $H(i,r)$, and let $H(i,r)=H(i,r-u_i^r-1)$.

\subsection{REWA Participant Selection Procedure}
Given the residual energy aware PS utility function and the corresponding REWA local computing policy, we summarize the REWAFL PS procedure in Algorithm~\ref{alg:REWAFL}.
Specifically, at the beginning of each FL training round,
each mobile device determines $H(i,r)$ according to Eqn.~(\ref{Eq:babeAdapH}) (Line 8). Given $H(i,r)$ of each mobile device, $|B_i^r|$, $Loss$, $t(i,r)$, $e(i,r)$ and $E_i^r$ can be calculated/estimated (Line 9).
Next, the mobile devices send those values to the edge server with negligible communication cost (Line 10).
When the edge server receives those values from mobile devices at the $r$-th round (Line 13), it calculates $Util(i,r)$ according to Eqn.~\ref{Eq:babeUtil} (Line 14).
Then, the edge server sorts $Util(i,r)$ in the descending order,
and selects the top $K$ devices for participation (Line 15).
After that, the edge server broadcasts the PS decision to mobile devices (Line 16).
After receiving the decision, if mobile device $i$ is selected to participate in the $r$-th round training,
it will update its residual energy (Line 20, i.e., $E_i^{r} \leftarrow E_i^{r-1} - e(i,r)$), the number of non-participating rounds (Line 21, $u_i^{r} \leftarrow 0$), and the number of local iterations (Line 22, $H(i,r) \leftarrow \left \lceil H(i,r-u_i^r-1) + \psi(s(i,r)) \Delta H \right \rceil$).
By contrast, if mobile device $i$ is not selected for participation in the $r$-th round,
$E_i^{r}$ and  $H(i,r)$ remain unchanged, while $u_i^{r}$ increases by 1 (Line 24-26).

\begin{algorithm}[th]
  \caption{REWAFL: Residual Energy and Wireless Aware Paticipant Selection.}
  \label{alg:REWAFL}
  \begin{algorithmic}[1]
    \State \textbf{Input:} Mobile device set $\mathcal{S}$, participant size $K$, weighting parameter $\alpha$ and $\beta$,  threshold energy reserve $E_0$, stop threshold $\varepsilon_{th}^r$.
    \State \textbf{Output:} Participants $\mathcal P$, Local iterations $\mathcal H$.
    \Statex {/* Initialize global variables. */}
    \State At the edge server, initialize participants $\mathcal P \leftarrow \emptyset$, training round $r \leftarrow 0$, and utility value of each mobile device $Util \leftarrow \emptyset$.
    \State At the mobile device, initialize the resdual energy $E^r \leftarrow E^0$, and the number of non-participating rounds $u^r \leftarrow 0$.
    \State For the $r$-th round FL training, where $r \in \{1,...,R\}$,
    \State \underline{\textbf{On Mobile Devices:}}
      \For{$i \in \mathcal{S}$}
         \Statex \ \ \quad  {/*Calculate  the number of local iterations. */}
         \State Calculate $H(i,r)$ according to Eqn.~(\ref{Eq:babeAdapH})
         \Statex \ \ \quad {/*Calculate/estimate  the utility parameters. */}
         \State Calculate/estimate $|B_i^r|$, $Loss$, $t(i,r)$, $e(i,r)$ and $E_i^r$ based on $H(i,r)$
         \State Send $|B_i^r|$, $Loss$, $t(i,r)$, $e(i,r)$ and $E_i^r$ to FL server
      \EndFor
      \State \underline{\textbf{On Edge Server:}}
      \State Receive the utility parameters from mobile devices
      \Statex {/*Calculate mobile device utility. */}
      \State Calculate $Util(i,r)$ according to Eqn.~(\ref{Eq:babeUtil})
      \Statex  {/*Sort mobile device utility and select participants.*/}
      \State $\mathcal P =$ RankingDevice($Util(i,r)$, K)
      \State Broadcast the device selection decision $\mathbb{V}(i,r)$
      \State \underline{\textbf{On Mobile Devices:}}
      \For{$i \in \mathcal{S}$}
          \If{$\mathbb{V}(i,r) = 1$}
          \Statex  \qquad \quad  {/*Update residual energy*/}
              \State $E_i^{r} \leftarrow E_i^{r-1}- e(i,r)$   
              \Statex  \qquad \quad  {/*Update number of non-participating rounds*/}
              \State $u_i^{r} \leftarrow 0$   
              \Statex  \qquad \quad  {/*Update number of local iterations*/}
              \State $H(i,r) \leftarrow \left \lceil H(i,r-u_i^r-1) + \psi(s(i,r)) \Delta H \right \rceil$    
          \Else
              \State $E_i^{r} \leftarrow E_i^{r-1}$  \Comment{Residual energy}
              \State $u_i^{r} \leftarrow u_i^{r-1}+1$  \Comment{Non-participating rounds}
              \State $H(i,r) \leftarrow  H(i,r-u_i^r-1)$ \Comment{Local iteration}
          \EndIf
      \EndFor

  \end{algorithmic}
\end{algorithm}
\vspace{-2mm}

\subsection{Self-Contained Staleness Solution}
\label{Sec:Staleness}
Since not every mobile device participates in every round of FL training, PS may lead to parameter staleness and biased training, and eventually degrade FL convergence or model accuracy. Most prior PS designs exploit a separated function to enforce those high-staleness mobile devices to participate and update their model parameters. For example, Oort~\cite{OortOSDI2021} forcibly increases the utility of long-neglected devices and makes them participate by using a ``temporal uncertainty'' mechanism, which is decoupled from Oort's original utility function design.

Different from existing PS designs, the proposed REWAFL addresses the staleness issue in a self-contained manner and buries the solution into the REWAFL's utility function and its local computing policy. Specifically, mobile devices with larger utility are more likely to be selected for training, where the utility of a mobile device is jointly determined by statistical utility, latency utility, and energy utility. As illustrated in Sec.~\ref{sec:AdaHPolicy}, given the training dataset, learning task and residual energy, a mobile device's statistical utility, latency utility and energy utility are all related to its local computing policy. Following the wireless aware local SGD computing strategy in Eqn.~(\ref{Eq:babeAdapH}), if mobile device $i$ is frequently selected to participate over rounds, its $H(i)$ continues increasing. Generally speaking, with $H(i)$'s increase over rounds, device $i$'s local training energy consumption increases and residual energy decreases, which leads to a decrease of its energy utility; its training latency increases, which leads to a decrease of its latency utility; its marginal contributions to FL's global model training are diminishing, which leads to a decrease in its statistical utility. The utility $Util(i)$ of frequently selected device $i$ keeps decreasing until it is less than that $Util(j)$ of a long-neglected device $j$, whose $H(j)$ has a small value, and then mobile device $j$ will be selected to participate in this training round. In this way, staleness issue can be addressed by REWAFL's inherent local SGD computing strategy (i.e., the wireless aware adaptive increasing of $H(\cdot)$) in a self-contained manner.

To further illustrate REWA's self-contained staleness solution, we take two mobile devices $i$ and $j$ with the same data distribution, computing capabilities and initial battery energy but different wireless transmission rates, say $s(i)> s(j)$, as a simple example.
At the beginning of FL training, since $s(i) > s(j)$, we have $H(i) < H(j)$ following Eqn.~(\ref{Eq:babeAdapH}), so that $e(i) < e(j)$ and $t(i) < t(j)$.
Given the fact that the local model updates of two mobile devices have similar contributions to learning at the beginning of FL training, we have $Util(i) > Util(j)$, and device $i$ will be selected for participation. As FL training
proceeds, $H(i)$ continuously increases over rounds following Eqn.~(\ref{Eq:babeAdapH}), while $H(j)$ remains unchanged. Until the $r$-th round, when $H(i,r)$ is larger than a certain value that triggers $Util(i,r) < Util(j,r)$, the staled mobile device $j$ will be selected to participate in this round of FL training. Similar analysis can be applied to mobile devices with data or/and device heterogeneity, where $H(\cdot)$ inherently serves as a tuning knob to tackle staleness in REWAFL.

\section{IMPLEMENTATION \& Experimental SETUP}
\label{SETUP}

\begin{figure}[!t]
\centering
\subfloat[]{\includegraphics[width=.6\linewidth]{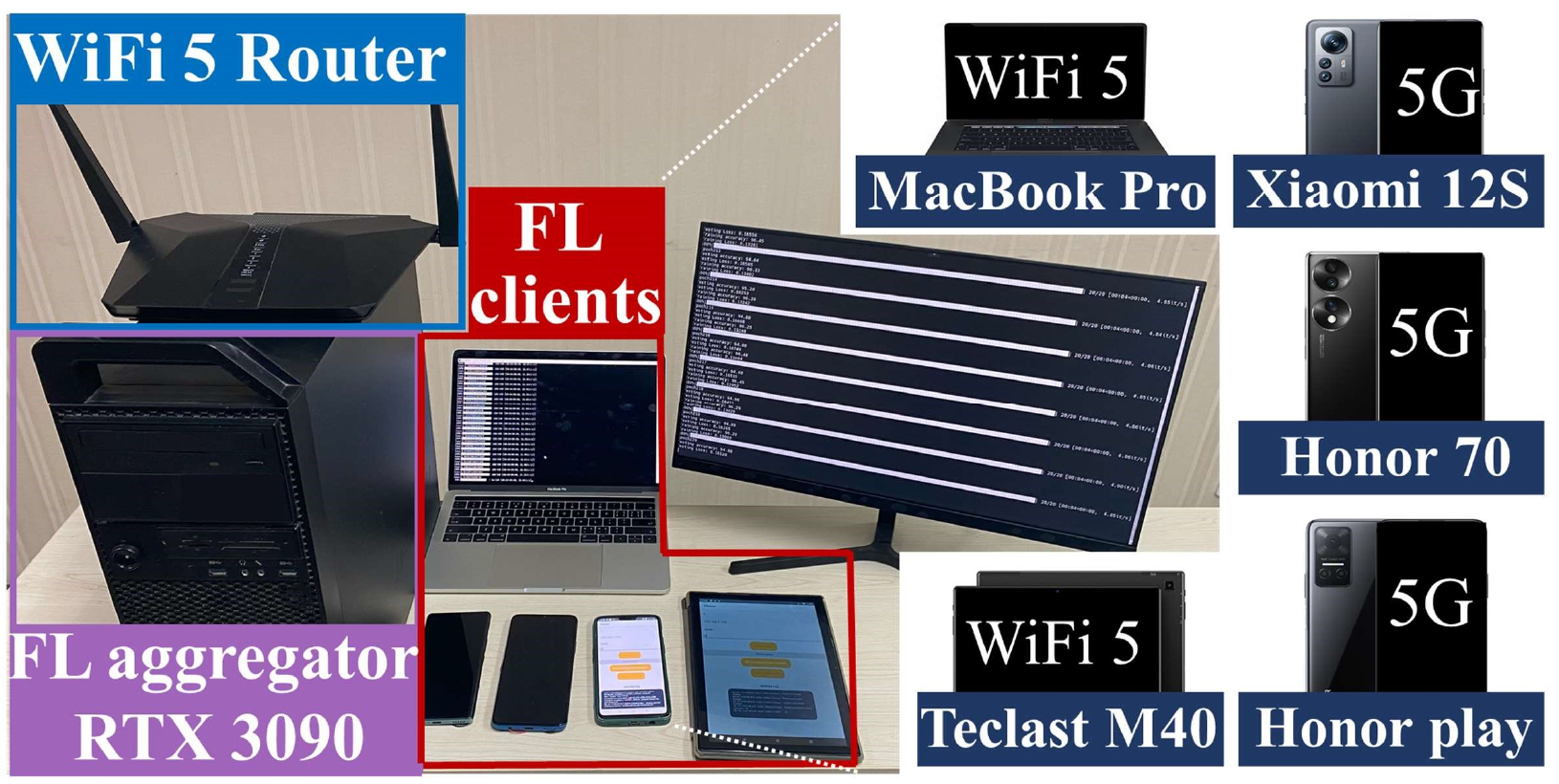}%
\label{Figure:testbed}}
\hfil
\subfloat[]{\includegraphics[width=.37\linewidth]{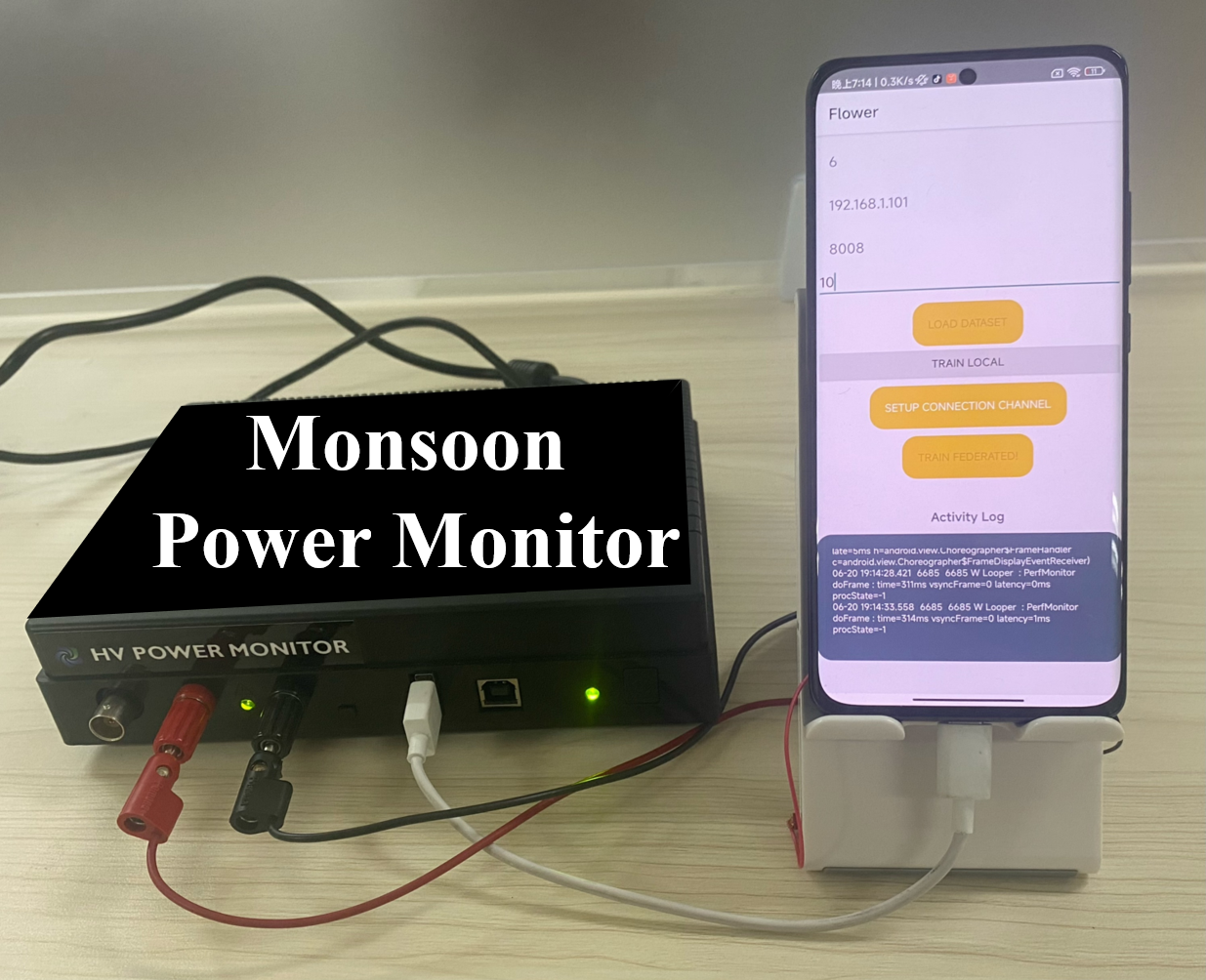}%
\label{Figure:Monitor}}
\caption{REWAFL testbed in the lab: (a) FL testbed configuration; (b) Xiaomi 12S' energy consumption measured by the Monsoon Power Monitor \cite{monsoon}.}
\label{Figure:Testbed}
\end{figure}

\subsection{REWAFL Implementation}
\label{IMPLEMENTATION}
We present the setup of testbed in Fig.~\ref{Figure:Testbed}.
Our REWAFL testbed consists of one model aggregation server and different types of mobile devices.
On FL aggregator side, we use a NVIDIA RTX 3090.
On FL client side, we consider five types of mobile devices:
(1) Xiaomi 12S smartphone with Qualcomm Snapdragon 8+ Gen 1 CPU, Adreno 730 GPU, 8GB RAM, and a battery capacity of 4500mAh;
(2) Honor 70 smartphone with Qualcomm Snapdragon 778G Plus CPU, Qualcomm Adreno 642L GPU, 12GB RAM, and a battery capacity of 5000mAh;
(3) Honor Play 6T smartphone with MediaTek Dimensity 700 CPU, Mali-G57 MC2 GPU, 8GB RAM, and a battery capacity of 5000mAh;
(4) Teclast M40 tablet with Unisoc Tiger T618 CPU, 6GB RAM, and a battery capacity of 7000mAh;
(5) MacBook Pro 2018 laptop with Intel Core i5-8259U CPU, Intel Iris Plus Graphics 655 GPU, 8GB RAM, and a battery capacity of 58Wh.
For each mobile device, its initial battery energy follows a normal distribution within the range of its battery's maximum capacity.
We set up a REWAFL system consisting of 100 mobile devices, with 20 devices for each type of mobile devices mentioned above.
We select 20 mobile devices for participation in every training round.

For communication between FL clients and FL aggregator, we employ a complex wireless transmission scenario, which consists of a mixture of Wi-Fi 5 and 5G networks.
Specifically,
the Teclast M40 tablet and MacBook Pro 2018 laptop are connected to FL aggregator via Wi-Fi 5, which uses the WebSocket communication protocol \cite{websoc}.  The Xiaomi 12S smartphone, Honor 70 smartphone, and Honor Play 6T smartphone are connected to FL aggregator via 5G  networks, which follows the 5G NR standard.
We further consider two types of transmission rates (high and low), and each device can be configured to one of these types.
As for a Wi-Fi transmission environment,
the command line tools nmcli and wondershaper are used for reporting
network status and controlling different wireless transmission rates.
As for 5G transmission environments,
we consider two transmission environments, i.e., indoor and outdoor, which represent different 5G transmission rates.


REWAFL is implemented by building on top of FLOWER \cite{beutel2021flower}. In total, we added a module to control the number of local iterations on the client side and a PS module on the server side, based on the original FLOWER code.
In order to realize REWAFL, we need to know the relevant parameters, such as transmission rate, training latency and energy consumption.
To estimate on-device transmission rates, we use the Network Monitor toollbox in the Android kernal on the smartphone.
To measure the latency, the time of model training and transmission are recorded on mobile devices.
To obtain the energy consumption, 
the Monsoon Power Monitor \cite{monsoon} is employed to measure the power consumption of smartphones and tablet.
As for laptop, HWiNFO is used to monitor the real-time power consumption.

\vspace{-2mm}
\subsection{Models, Datasets and Parameters}
\label{Datasets}
We evaluate REWAFL' performance for three different FL tasks: image classification,  next word prediction, and human activity recognition.
We consider two DNN models: 2-layer CNN \cite{pmlr-v54-mcmahan17a} and  LSTM \cite{hochreiter1997long}.

As for the image classification task, we use two datasets.
One is a MNIST dataset \cite{deng2012mnist} consisting of 10 categories ranging from digit ``0'' to ``9'', which includes 60,000 training images and 10,000 validation images.
The other is CIFAR10 dataset \cite{krizhevsky2009learning} consisting of 10 categories, which includes 50000 training images and 10,000 validation images.
As for data distribution among mobile devices,
we denote $\lambda$ as the non-i.i.d. levels, where $\lambda=0$  indicates that the data among mobile devices is i.i.d., $\lambda=0.8$ indicates that 80\% of the data belong to one label and the remaining 20\% data belong to other labels, and $\lambda=1$ indicates that each mobile device owns a disjoint subset of data with one label.
We train the image classification data samples with 2-layer CNN.

As for next word prediction task, we use Shakespeare dataset \cite{Shakespeare} that consists of 1,129 roles and each of which is viewed as a device. Since the number of lines and speaking habits of each role vary greatly, the dataset is non-i.i.d. and unbalanced.
We train the Shakespeare data samples with LSTM.

As for task of human activity recognition,
we focus on a publicly accessible dataset generated by having volunteers wear Samsung Galaxy S2 smartphones equipped with accelerometer and gyroscope sensors \cite{Stisen2015SmartDA}.
The dataset includes six categories of activities, i.e., walking, walking upstairs, walking downstairs, sitting, standing and laying. The dataset is non-i.i.d. due to the different behavioral habits of the volunteers.
It contains 10,299 data samples.
We employ a 2-layer CNN to train it.

The default parameter settings are given as follows unless specified otherwise.
For the MNIST and CIFAR10 datasets, the non-i.i.d. level of data distribution among mobile devices is set as $\lambda = 0.8$.
Besides, the scaling coefficients in the proposed REA PS utility function are set as $\alpha = 1$ and $\beta = 1$.

\subsection{Baselines for Comparison}
\label{Baselines}
We compared REWAFL with the following peer FL designs under different learning tasks, DNN models and datasets.

Random \cite{Yang20FedAVG}: The FL server selects participants randomly, and the selected mobile devices utilize a fixed local computing policy.

Oort \cite{OortOSDI2021}: The FL server selects mobile devices based on global FL training utility function as defined in Eqn.~(\ref{Eq:oortUtil}). The selected mobile devices utilize a fixed local computing policy.

AutoFL \cite{kim2021autofl}: The FL server selects mobile devices based on mobile devices' energy consumption, and the selected mobile devices utilize a fixed local computing policy.

REAFL: The FL server selects mobile devices based on REA PS utility function as defined in Eqn.~(\ref{Eq:babeUtil}),
and the selected mobile devices utilize a fixed local computing policy.

REAFL+LUPA: The FL server selects mobile devices based on REA PS utility function as defined in Eqn.~(\ref{Eq:babeUtil}), and the selected mobile devices  utilize the AdaH strategy during FL training \cite{Haddadpour19AdaH}, i.e.,
$H(i,r)= \left \lceil H(0) + \sum\limits_{l = 1}^{r} \psi\Delta H \right \rceil$.

\begin{table*}[]
   \caption{Performance Comparison: Dropout Ratio (DR), Overall Latency (OL) and Overall Energy Consumption (OEC) of FL Training with Non-i.i.d. Data to Reach the Target Testing Accuracy.
   \label{table:comparisonSOTA}}
\small
\resizebox{\textwidth}{19mm}{
\begin{tabular}{@{}ccccccccccccc@{}}
\toprule
Task Type   & \multicolumn{6}{c}{CV}& \multicolumn{3}{c}{NLP}& \multicolumn{3}{c}{HAR} \\ \cmidrule(lr){2-7} \cmidrule(lr){8-10} \cmidrule(lr){11-13}
Local Model & \multicolumn{3}{c}{CNN@MNIST} & \multicolumn{3}{c}{CNN@CIFAR10}  & \multicolumn{3}{c}{LSTM@Shakespeare}     & \multicolumn{3}{c}{CNN@HAR} \\\midrule
Target Accuracy & \multicolumn{3}{c}{91.0\%} & \multicolumn{3}{c}{72.2\%} &\multicolumn{3}{c}{50.3\%} &\multicolumn{3}{c}{89.3\%} \\\midrule
Methods & DR (\%) & OL (h) & OEC (kJ) & DR (\%) & OL (h) & OEC (kJ) & DR (\%) & OL (h) & OEC (kJ) & DR (\%) & OL (h) & OEC (kJ)  \\\midrule
Random
& 7.0 & 4.9
& 1137.5 & 18.0
& 22.4 & 4349.5
& 38.0 & 32.8
& 5031.2 & 28.0
& 6.5 & 1538.4          \\
Oort
& 46.0 & 4.1
& 1230.3 & 46.0
& 13.1 & 4750.5
& 66.0 & 19.0
& 5217.1 & 85.0
& 5.6 & 1561.9 \\
AutoFL
& 37.0 & 8.1
& 1069.5 & 25.0
& 28.7 & 3911.9
& 53.0 & 34.1
& 4835.4 & 55.0
& 9.1 & 1498.6          \\
REAFL
& \textbf{0.0} & \textbf{2.0}
& \textbf{562.8} & \textbf{0.0}
& \textbf{7.2} & \textbf{3613.6}
& \textbf{2.0} & \textbf{11.8}
& \textbf{4267.7} & \textbf{0.0}
& \textbf{3.0} & \textbf{704.7}
\\ \bottomrule
\end{tabular}}
\vspace{-2mm}
\end{table*}

\section{EVALUATION RESULTS AND ANALYSIS}

\subsection{Advantages of REA PS Utility Function
}

\begin{figure}[!t]
\centering
\subfloat[]{\includegraphics[width=.49\linewidth]{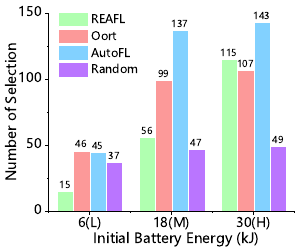}%
\label{Figure:selection_num_high}}
\hfil
\subfloat[]{\includegraphics[width=.49\linewidth]{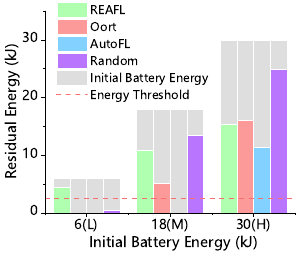}%
\label{Figure:Eres_high}}
\caption{Performance comparison of different PS utility designs (CNN@MNIST, Xiaomi 12S smartphone, 5G-79.60Mbps). (a) the number of selections, and (b) energy consumption and residual energy.}
\label{Figure:PSfuncHighEndDevice}
\end{figure}

\subsubsection{Awareness of devices' heterogeneous residual energy}
As shown in Table~\ref{table:comparisonSOTA}, given a targeted testing accuracy, the dropout ratio of \emph{REAFL} is much lower than that of \emph{Random}, \emph{Oort}, or \emph{AutoFL} for all learning tasks. The reason is simple. \emph{REAFL} employs the proposed residual energy aware PS utility function and avoids frequently selecting mobile devices with low residual energy, while other SOTA PS designs are not aware of participating mobile devices' residual energy levels at all. That may result in the depletion of energy for frequently selected high-end mobile devices (i.e., mobile devices with good learning contribution, short latency, and/or small energy consumption), especially for those mobile devices with low initial residual energy.

To further verify the above analysis, we have recorded the number of selections and energy consumption for the high-end mobile devices (Xiaomi 12S smartphone with the averaged 5G uplink transmission rate of 79.60Mbps) with heterogeneous initial residual energy levels (i.e., 6kJ--low, 18kJ--medium, and 30kJ--high) under different PS utility function designs, and shown the results in Fig.~\ref{Figure:PSfuncHighEndDevice}. From Fig.~\ref{Figure:PSfuncHighEndDevice}(a) and Fig.~\ref{Figure:PSfuncHighEndDevice}(b), we find that the proposed REA PS utility function represented by \emph{REAFL} is aware of heterogeneous residual energy among mobile devices, so it doesn't select the high-end mobile device with low residual energy too often, and keeps the necessary energy for participating devices' mandatory operations. By contrast, other PS utility function designs (i.e., \emph{Random}, \emph{Oort}, or \emph{AutoFL}) are not aware of mobile devices' residual energy, and often select high-end mobile devices to participate in training even when their residual
energy is low, thus may drain out the battery of
such devices.

\subsubsection{REAFL energy and latency efficiency analysis}
Besides cutting off dropout, \emph{REAFL} also helps to reduce the overall energy consumption and latency of FL training to achieve a target testing accuracy. As shown in Table~\ref{table:comparisonSOTA}, in terms of overall latency, \emph{REAFL} reduces around 51.2\%, 45.0\%, 37.9\% and 46.4\% for learning tasks CNN@MNIST, CNN@CIFAR10, LSTM@Shakespeare, and CNN@HAR, compared with \emph{Oort}, respectively, around 75.3\%, 74.9\%, 65.4\% and 67.0\% compared with \emph{AutoFL},
and around 59.2\%, 67.9\%, 64.0\% and 53.8\% compared with \emph{Random};
in terms of overall energy consumption, \emph{REAFL} saves around 54.3\%, 23.9\%, 18.2\% and 54.9\% for learning tasks CNN@MNIST, CNN@CIFAR10, LSTM@Shakespeare, and CNN@HAR, compared with \emph{Oort},
around 47.4\%, 7.6\%, 11.7\% and 53.0\% compared with \emph{AutoFL},
and around 50.5\%, 16.9\%, 15.2\% and 54.2\% compared with \emph{Random}.
The rationale behind it is that SOTA PS utility function designs (i.e., \emph{Random}, \emph{Oort}, or \emph{AutoFL}) are not aware of mobile devices' residual energy, and thus frequently select the mobile devices with good learning contributions and small latency/energy consumption at early training stages, which may drain up their energy. Thus, at the late stages of FL training, those depleted mobile devices are not able to participate. Their absence will slow down FL convergence, in particular for non-i.i.d. cases, and thus increase overall latency/energy consumption.

\begin{table*}[th]
   \caption{Summary of REWAFL' Performance Improvement over REAFL and REAFL + LUPA
   \label{table:comparisonwREWAFL}}
   \centering
\small
\begin{tabular}{@{}ccccccccccccc@{}}
\toprule
Task  Type  & \multicolumn{4}{c}{CV}& \multicolumn{2}{c}{NLP}& \multicolumn{2}{c}{HAR} \\ \cmidrule(lr){2-5} \cmidrule(lr){6-7} \cmidrule(lr){8-9}
Local Model & \multicolumn{2}{c}{CNN@MNIST} & \multicolumn{2}{c}{CNN@CIFAR10}  & \multicolumn{2}{c}{LSTM@Shakespeare}     & \multicolumn{2}{c}{CNN@HAR} \\\midrule
Target Accuracy & \multicolumn{2}{c}{91.0\%} & \multicolumn{2}{c}{72.2\%} &\multicolumn{2}{c}{50.3\%} &\multicolumn{2}{c}{89.3\%} \\\midrule
Methods  & OL (h) & OEC (kJ) & OL (h) & OEC (kJ)   & OL (h) & OEC (kJ)   & OL (h) & OEC (kJ)   \\\midrule
REAFL
& 2.0 & 562.8
& 7.2 & 3613.6
& 11.8 & 4267.7
& 3.0  & 704.7
\\
REAFL+LUPA
& 1.7   & 473.6
& 7.0       & 3503.7
& 11.4  & 4004.8
& 2.3   & 603.5  \\
REWAFL
& \textbf{1.3}   & \textbf{357.6}
& \textbf{6.8}       & \textbf{3203.2}
& \textbf{10.9}  & \textbf{3716.1}
& \textbf{2.1}   & \textbf{569.7}
\\ \bottomrule
\end{tabular}
\vspace{-2mm}
\end{table*}

\subsection{Further Performance Improvement by the REWA Local Computing Policy}
Under the proposed \emph{REAFL} PS utility function framework,
we further evaluate the potential performance improvement brought by REWA local computing policy in terms of overall latency and overall energy consumption for FL training.

As shown in Table~\ref{table:comparisonwREWAFL}, compared with \emph{REAFL} (i.e., the best PS design in last subsection), when FL training reaches the target accuracy, \emph{REWAFL} reduces around 35.0\%, 5.6\%, 7.6\% and 30.0\% latency for learning tasks CNN@MNIST, CNN@CIFAR10, LSTM@Shakespeare and CNN@HAR, and 36.5\%,  11.4\%, 12.9\% and 19.2\% energy consumption. Different from \emph{REAFL}'s fixed local computing policy, the performance improvement comes from \emph{REWAFL}'s REWA local computing policy, which increases global statistically utility, latency utility, and local device's energy utility, and helps to speed up convergence and reduce overall latency/energy consumption.

Compared with \emph{REAFL+LUPA}, when FL model reaches target accuracy, \emph{REWAFL} reduces around 23.5\%, 2.9\%, 4.4\% and 8.7\% latency for learning tasks CNN@MNIST, CNN@CIFAR10, LSTM@Shakespeare and CNN@HAR, and 24.5\%,  8.6\%, 7.2\% and 5.6\% energy consumption. Although \emph{REAFL+LUPA} employs the AdaH policy~\cite{Haddadpour19AdaH}, its performance is not as good as \emph{REWAFL} because \emph{REAFL+LUPA}'s local computing policy fails to consider the impacts of devices' heterogeneous wireless transmissions, and ignores the trade-off between learning performance benefits/statistical utility and the energy cost/energy utility of increasing $H$. Purely reducing the number of communication rounds while keeping increasing $H$ without stopping criteria will inevitably increase the overall latency/energy consumption. By contrast, \emph{REWAFL} has awareness of wireless transmission heterogeneity and develops an energy utility aware stopping criterion for increasing $H$, and exploit them to improve the latency/energy efficiency of FL training over mobile devices.

\subsection{
REWAFL's Staleness Analysis
}
\begin{figure}[!t]
\centering
\subfloat[]{\includegraphics[width=0.49\linewidth]{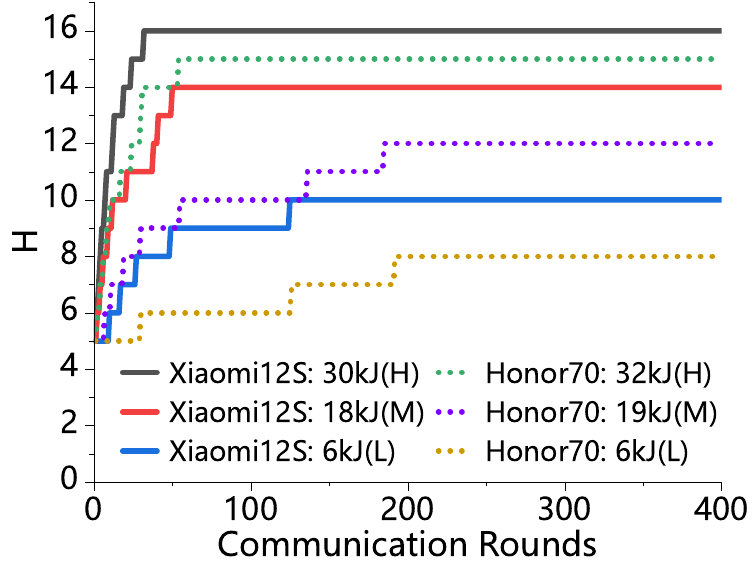}
\label{Figure:HE}}
\hfil
\subfloat[]{\includegraphics[width=0.49\linewidth]{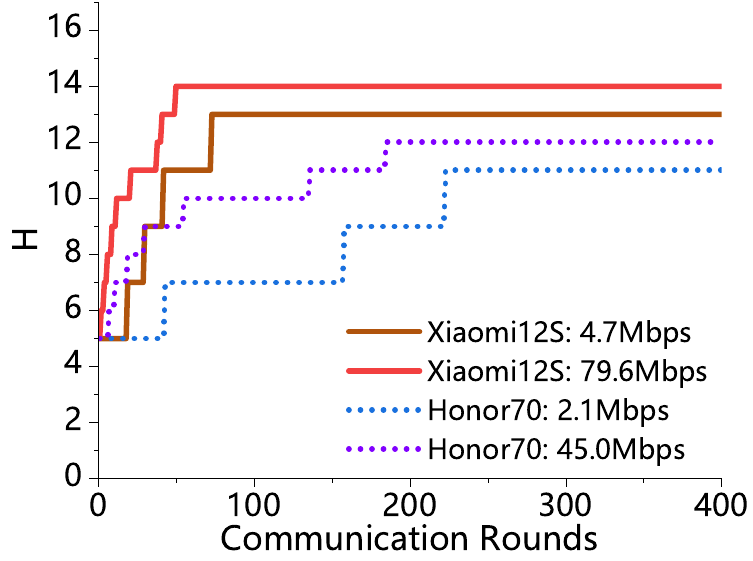}%
\label{Figure:Hcomm}}
\caption{REWAFL's staleness solution. (a) different initial energy; (b) different tx. rates (CNN@MNIST).}
\label{Figure:H}
\end{figure}

To analyze \emph{REWAFL}'s self-contained staleness solution, we employ two types of mobile devices: Xiaomi 12S and Honor 70, representing high-end and low-end mobile devices, respectively. Figure~\ref{Figure:H} presents the changes of $H$ with high-end and low-end mobile devices, which shows the growth frequency, increment of $H$ and the final stopping value.
According to Eqn.~(\ref{Eq:babeAdapH}), $H$ increases only if the device is selected, and the increment of $H$ is determined by the wireless aware local computing policy. Besides, the saturated value of $H$ is determined by Eqn.~(\ref{Eq:AdapHstop}).

Figure~\ref{Figure:H}(a) shows the changes of $H$ for mobile devices with different initial battery energy levels (i.e., high, medium and low). Here, we configure the same transmission rate for the same type of mobile devices, i.e., Xiaomi 12S with the averaged 5G uplink transmission rate of 79.6 Mbps, and Honor 70 smartphones with the averaged 5G uplink transmission rate of 45.0 Mbps.
As shown in the figure, for the same type of mobile devices, the $H$'s saturated values of the mobile devices with high initial energy are larger than those of devices with low initial energy. That is consistent with Eqn.~(\ref{Eq:AdapHstop}) and the energy utility aware stopping criterion analysis in Sec.~\ref{sec:stoppingH}. Moreover, as for different types of mobile devices with the same/similar level of initial battery energy, $H$' growth frequency is high for high-end mobile devices (i.e., Xiaomi 12S) at the beginning of the training process.
It is because high-end mobile devices have low energy consumption/latency, leading to frequent selections of these mobile devices for participation. However, in the later stages of training, the $H$ growth frequency of low-end mobile devices (i.e., Honor 70) is higher than that of high-end mobile devices. Similar to the analysis in Sec.~\ref{Sec:Staleness}, as the high-end mobile device is consecutively selected, its $H$ increases and PS utility keeps decreasing until it is less than that of the low-end mobile device. Then, the low-end mobile device will be selected to participate in the FL training. The results demonstrate that the proposed \emph{REWAFL} addresses the staleness issue in a self-contained manner.


Figure~\ref{Figure:H}(b) presents the changes of $H$ for mobile devices with different wireless transmission rates.
First, we have similar observations as those in Fig.~\ref{Figure:H}(a) for high-end and low-end devices. Besides, for the same type of devices, high transmission rate leads to a small $H$ increment in each communication round, and a high growth frequency of $H$ in the early stages of training. That is because \emph{REWAFL} employs a wireless aware local SGD computing strategy, which makes mobile devices with  higher transmission rate have a smaller increase in $H$, so that its computing latency and energy consumption can be reduced. Thus, the mobile devices with higher transmission rate has a better chance to be selected, which increases their growth frequency of $H$ at the early stages of training. However, the utility of devices with higher transmission rate decreases as
$H$ increases. Once it falls below the utility of mobile devices with lower transmission rate, the mobile device with low transmission rate is then selected to participate. Therefore, in the later stages of training, $H$ of the mobile devices with lower transmission rate grows due to being selected. This is consistent with our analysis of \emph{REWAFL}'s wireless aware local computing policy in Sec.~\ref{sec:AdaHPolicy} and self-contained staleness solution in Sec.~\ref{Sec:Staleness}.

\begin{table*}[th]
   \caption{Performance Evaluation under Data Heterogeneity.\label{table:comparison}}
   \centering
\small
\begin{tabular}{@{}ccccccccccccc@{}}
\toprule
Local Model & \multicolumn{9}{c}{CNN@MNIST}  \\\midrule
Data Heterog.   & \multicolumn{3}{c}{$\lambda = 0$}& \multicolumn{3}{c}{$\lambda = 0.8$}& \multicolumn{3}{c}{$\lambda = 1$} \\ \cmidrule(lr){2-4} \cmidrule(lr){5-7} \cmidrule(lr){8-10}
Target Accuracy & \multicolumn{3}{c}{97.0\%} & \multicolumn{3}{c}{91.0\%} &\multicolumn{3}{c}{89.9\%}  \\\midrule
Methods  & OL (h) & OEC (kJ) & DR (\%)  & OL (h) & OEC (kJ) & DR (\%)  & OL (h) & OEC (kJ) & DR (\%)\\\midrule
Random
 & 3.1
& 803.8 & 9.0
 & 4.9
& 1137.5 & 7.0
 & 5.0
& 1215.6 & 25.0\\
Oort
 & \textbf{1.3}
& 650.1 & 24.0
 & 4.1
& 1230.3 & 46.0
 & 5.0
& 1451.8 & 48.0\\
AutoFL
 & 1.6
& \textbf{411.2} & 5.0
 & 8.1
& 1069.5 & 37.0
 & 8.7
& 1118.2 & 38.0 \\
REWAFL
 & 2.2
& 624.9 & \textbf{0.0}
& \textbf{1.3}
& \textbf{357.6} & \textbf{0.0}
 & \textbf{2.6}
& \textbf{650.1} & \textbf{0.0}
\\ \bottomrule
\end{tabular}
\vspace{-2mm}
\end{table*}


\begin{figure}[!t]
\centering
\subfloat[]{\includegraphics[width=.49\linewidth]{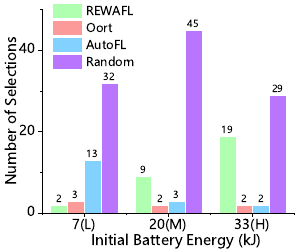}%
\label{Figure:selection_num_low}}
\hfil
\subfloat[]{\includegraphics[width=.49\linewidth]{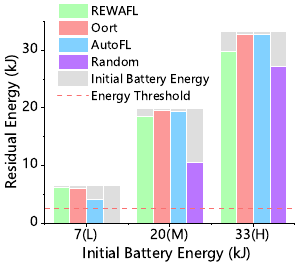}%
\label{{Figure:Eres_low}}}
\caption{Performance comparison of staleness solutions in different PS approaches (CNN@MNIST, Honor Play 6T, 5G-0.64Mbps). (a) the number of selections, and (b) energy consumption and residual energy.}
\label{Figure:zhuzhuangtu2}
\end{figure}

Then, we demonstrate the advantages of \emph{REWAFL}'s staleness solution by comparing the number of selections and residual energy for the low-end mobile devices (i.e., Honor Play 6T smartphones) with low transmission rates (i.e., averaged 5G uplink transmission rate of 0.64Mbps) under different PS designs.
The results are shown in Fig.~\ref{Figure:zhuzhuangtu2}.
Compared with \emph{Random}, \emph{Oort} and \emph{AutoFL}, the low-end mobile devices have a better chance to be selected for participation in training without depleting their energy in \emph{REWAFL}.
It indicates that \emph{REWAFL} can address the staleness issue in a self-contained manner while outperforming SOTA PS designs and/or their isolated staleness solutions.

\begin{figure*}[!t]
\centering
\subfloat[]{\includegraphics[width=.28\linewidth]{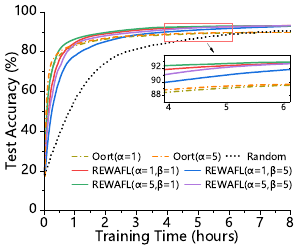}%
\label{Figure:sensitivity_time}}
\hfil
\subfloat[]{\includegraphics[width=.28\linewidth]{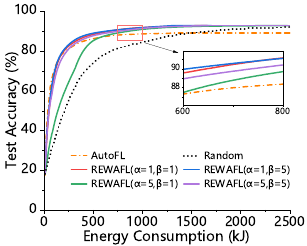}%
\label{Figure:sensitivity_energy}}
\hfil
\subfloat[]{\includegraphics[width=.28\linewidth]{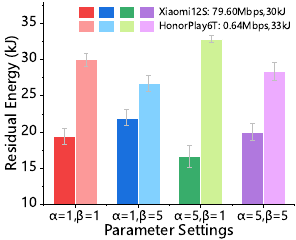}%
\label{Figure:sensitivity_Resenergy}}
\caption{REWAFL's learning performance and efficiency with different $\alpha$ and $\beta$ values (CNN@HAR, $\lambda$ = 0.8). (a) Test accuracy v.s. training time, (b) Test accuracy v.s. energy consumption, and (c) Residual energy.}
\label{Figure:sensitivity}
\end{figure*}

\subsection{Sensitivity Analysis}
To analyze \emph{REWAFL}'s sensitivity to different parameter settings, we further evaluate the performance of \emph{REWAFL} with different $\alpha$ and $\beta$ values. Figure~\ref{Figure:sensitivity} shows that \emph{REWAFL} outperforms its counterparts across different $\alpha$ and $\beta$ values. As shown in Fig.~\ref{Figure:sensitivity}(a), a larger $\alpha$ makes the system focus more on latency utility, which allows the global model to reach the same learning performance in a smaller latency. Correspondingly, a larger $\beta$ makes the system focus more on improving the energy utility, and allows the global model to reach the same test accuracy with less energy consumption as shown in Fig.~\ref{Figure:sensitivity}(b).
Besides, as shown in Fig.~\ref{Figure:sensitivity}(c),
a larger $\beta$ leaves more residual energy for the frequently selected high-end devices (i.e., Xiaomi 12S) and less residual energy for the infrequently selected low-end devices (i.e., Honor Play 6T), which
balances the energy consumption among participating mobile devices.
Figure~\ref{Figure:sensitivity} indicates that $\alpha$ and $\beta$ can serve as scaling coefficients to associate/balance different utilities in terms of accuracy, latency and energy efficiency.

\vspace{-3mm}
\subsection{Impacts of Data Heterogeneity}
We further evaluate the performance of \emph{REWAFL} with different non-i.i.d levels of training data samples in CNN@MNIST case.
The results are also applicable to other models and datasets.
As shown in Table~\ref{table:comparison},  \emph{REWAFL} has the best  performance  in terms of dropout ratio.
For non-i.i.d. cases (i.e., $\lambda = 0.8$ and $\lambda = 1$), \emph{REWAFL} has a better overall latency and overall energy consumption to reach the target accuracy.
For i.i.d. data distribution case (i.e., $\lambda = 0$), we find that \emph{REWAFL} has similar performance to \emph{Oort} in terms of overall latency to reach the target accuracy, and similar performance to \emph{AutoFL} in terms of overall energy consumption.
The potential reason is that the impact of device dropouts on convergence rate is not significant due to the similar contribution of local data from different devices to the global model. Unsurprisingly, even for i.i.d. case, \emph{REWAFL} can still maintain the lowest dropout ratio, which reserves enough residual energy at mobile devices for their mandatory operations while enjoying the global learning results.
Besides, we find that the FL system with non-i.i.d. data has a higher latency, energy consumption and dropout ratio than that with i.i.d. data to achieve the same accuracy goal.

\vspace{-3mm}
\section{CONCLUSION}
In this paper, we have developed a residual energy and wireless aware PS design for efficient FL training over mobile devices (REWAFL). We have observed that most existing participation  selection approaches are (i) unaware of heterogeneous residual battery energy, and (ii) unaware of heterogeneous wireless transmission rates among candidate/participating mobile devices.
To address these challenges, REWAFL introduces a novel PS utility function considering statistical utility, global latency utility and local energy utility. Under the proposed PS utility function framework, REWAFL further provides a residual energy and wireless aware local computing policy to improve FL efficiency. Moreover, REWAFL buries the staleness solution into its utility function and local computing policy development without incurring any additional mechanisms. Extensive experimental results have demonstrated the effectiveness of the proposed REWAFL, and its efficiency superiority over peer PS designs under various learning tasks, DNN models, datasets for FL training over mobile devices.




\bibliographystyle{IEEEtran}
\bibliography{li_ref}

\begin{thebibliography}{10}
\providecommand{\url}[1]{#1}
\csname url@samestyle\endcsname
\providecommand{\newblock}{\relax}
\providecommand{\bibinfo}[2]{#2}
\providecommand{\BIBentrySTDinterwordspacing}{\spaceskip=0pt\relax}
\providecommand{\BIBentryALTinterwordstretchfactor}{4}
\providecommand{\BIBentryALTinterwordspacing}{\spaceskip=\fontdimen2\font plus
\BIBentryALTinterwordstretchfactor\fontdimen3\font minus
  \fontdimen4\font\relax}
\providecommand{\BIBforeignlanguage}[2]{{%
\expandafter\ifx\csname l@#1\endcsname\relax
\typeout{** WARNING: IEEEtran.bst: No hyphenation pattern has been}%
\typeout{** loaded for the language `#1'. Using the pattern for}%
\typeout{** the default language instead.}%
\else
\language=\csname l@#1\endcsname
\fi
#2}}
\providecommand{\BIBdecl}{\relax}
\BIBdecl

\bibitem{Pan21infocom}
L.~Li, D.~Shi, R.~Hou, H.~Li, M.~Pan, and Z.~Han, ``To talk or to work:
  Flexible communication compression for energy efficient federated learning
  over heterogeneous mobile edge devices,'' in \emph{IEEE INFOCOM 2021 - IEEE
  Conference on Computer Communications}, 2021, pp. 1--10.

\bibitem{Zhu20MCOM}
G.~Zhu, D.~Liu, Y.~Du, C.~You, J.~Zhang, and K.~Huang, ``Toward an intelligent
  edge: Wireless communication meets machine learning,'' \emph{IEEE
  Communications Magazine}, vol.~58, no.~1, pp. 19--25, 2020.

\bibitem{Qin22JSAC}
W.~Guo, R.~Li, C.~Huang, X.~Qin, K.~Shen, and W.~Zhang, ``Joint device
  selection and power control for wireless federated learning,'' \emph{IEEE
  Journal on Selected Areas in Communications}, vol.~40, no.~8, pp. 2395--2410,
  2022.

\bibitem{Cho2020ClientSI}
Y.~J. Cho, J.~Wang, and G.~Joshi, ``Client selection in federated learning:
  Convergence analysis and power-of-choice selection strategies,''
  \emph{ArXiv}, vol. abs/2010.01243, 2020.

\bibitem{2017LearningWP}
A.~D.~P. Team, ``Learning with privacy at scale differential,'' in \emph{Apple
  Machine Learning Journal}, 2017.

\bibitem{geyer2019differentially}
\BIBentryALTinterwordspacing
R.~C. Geyer, T.~J. Klein, and M.~Nabi, ``Differentially private federated
  learning: A client level perspective,'' 2019. [Online]. Available:
  \url{https://openreview.net/forum?id=SkVRTj0cYQ}
\BIBentrySTDinterwordspacing

\bibitem{oort34}
F.~{Hartmann}, S.~{Suh}, A.~{Komarzewski}, T.~D. {Smith}, and I.~{Segall},
  ``{Federated Learning for Ranking Browser History Suggestions},'' \emph{arXiv
  e-prints}, Nov. 2019.

\bibitem{Fan2020Sol}
F.~Lai, J.~You, X.~Zhu, H.~V. Madhyastha, and M.~Chowdhury, ``Sol: Fast
  distributed computation over slow networks,'' in \emph{17th USENIX Symposium
  on Networked Systems Design and Implementation (NSDI 20)}.\hskip 1em plus
  0.5em minus 0.4em\relax Santa Clara, CA: USENIX Association, Feb. 2020, pp.
  273--288.

\bibitem{li2020federated}
T.~Li, A.~K. Sahu, A.~Talwalkar, and V.~Smith, ``Federated learning:
  Challenges, methods, and future directions,'' \emph{IEEE Signal Processing
  Magazine}, vol.~37, no.~3, pp. 50--60, 2020.

\bibitem{nguyen2021federated}
D.~C. Nguyen, M.~Ding, P.~N. Pathirana, A.~Seneviratne, J.~Li, and H.~V. Poor,
  ``Federated learning for internet of things: A comprehensive survey,''
  \emph{arXiv preprint arXiv:2104.07914}, 2021.

\bibitem{Ye20}
D.~Ye, R.~Yu, M.~Pan, and Z.~Han, ``Federated learning in vehicular edge
  computing: A selective model aggregation approach,'' \emph{IEEE Access},
  vol.~8, pp. 23\,920--23\,935, 2020.

\bibitem{OortOSDI2021}
F.~Lai, X.~Zhu, H.~V. Madhyastha, and M.~Chowdhury, ``Oort: Efficient federated
  learning via guided participant selection,'' in \emph{15th {USENIX} Symposium
  on Operating Systems Design and Implementation ({OSDI} 21)}.\hskip 1em plus
  0.5em minus 0.4em\relax {USENIX} Association, Jul. 2021, pp. 19--35.

\bibitem{Ren20Scheduling}
J.~Ren, Y.~He, D.~Wen, G.~Yu, K.~Huang, and D.~Guo, ``Scheduling for cellular
  federated edge learning with importance and channel awareness,'' \emph{IEEE
  Transactions on Wireless Communications}, vol.~19, no.~11, pp. 7690--7703,
  2020.

\bibitem{Tianyi18NIPS}
T.~Chen, G.~B. Giannakis, T.~Sun, and W.~Yin, ``Lag: Lazily aggregated gradient
  for communication-efficient distributed learning,'' in \emph{Proceedings of
  the 32nd International Conference on Neural Information Processing Systems},
  ser. NIPS'18.\hskip 1em plus 0.5em minus 0.4em\relax Red Hook, NY, USA:
  Curran Associates Inc., 2018, p. 5055–5065.

\bibitem{Katharopoulos2018NotAS}
A.~Katharopoulos and F.~Fleuret, ``Not all samples are created equal: Deep
  learning with importance sampling,'' in \emph{International Conference on
  Machine Learning}, 2018.

\bibitem{Salehi18Sampling}
F.~Salehi, P.~Thiran, and L.~E. Celis, ``Coordinate descent with bandit
  sampling,'' in \emph{Proceedings of the 32nd International Conference on
  Neural Information Processing Systems}, ser. NIPS'18.\hskip 1em plus 0.5em
  minus 0.4em\relax Red Hook, NY, USA: Curran Associates Inc., 2018, p.
  9267–9277.

\bibitem{Wu22noniid}
H.~Wu and P.~Wang, ``Node selection toward faster convergence for federated
  learning on non-iid data,'' \emph{IEEE Transactions on Network Science and
  Engineering}, vol.~9, no.~5, pp. 3099--3111, 2022.

\bibitem{Zhu20Analog}
G.~Zhu, Y.~Wang, and K.~Huang, ``Broadband analog aggregation for low-latency
  federated edge learning,'' \emph{IEEE Transactions on Wireless
  Communications}, vol.~19, no.~1, pp. 491--506, 2020.

\bibitem{Amiri2021Scheduling}
M.~M. Amiri, D.~Gündüz, S.~R. Kulkarni, and H.~V. Poor, ``Convergence of
  update aware device scheduling for federated learning at the wireless edge,''
  \emph{IEEE Transactions on Wireless Communications}, vol.~20, no.~6, pp.
  3643--3658, 2021.

\bibitem{kim2021autofl}
Y.~G. Kim and C.-J. Wu, ``Autofl: Enabling heterogeneity-aware energy efficient
  federated learning,'' in \emph{MICRO-54: 54th Annual IEEE/ACM International
  Symposium on Microarchitecture}, 2021, pp. 183--198.

\bibitem{RuiEEFL}
R.~Chen, Q.~Wan, X.~Zhang, X.~Qin, Y.~Hou, D.~Wang, X.~Fu, and M.~Pan,
  ``{EEFL}: High-speed wireless communications inspired energy efficient
  federated learning over mobile devices,'' in \emph{Proceedings of the 21st
  Annual International Conference on Mobile Systems, Applications and
  Services}, ser. MobiSys '23.\hskip 1em plus 0.5em minus 0.4em\relax New York,
  NY, USA: Association for Computing Machinery, 2023, p. 544–556.

\bibitem{lyx22TVT}
Y.~Li, X.~Qin, H.~Chen, K.~Han, and P.~Zhang, ``Energy-aware edge association
  for cluster-based personalized federated learning,'' \emph{IEEE Transactions
  on Vehicular Technology}, vol.~71, no.~6, pp. 6756--6761, 2022.

\bibitem{Haddadpour19AdaH}
F.~Haddadpour, M.~M. Kamani, M.~Mahdavi, and V.~R. Cadambe, \emph{Local SGD
  with Periodic Averaging: Tighter Analysis and Adaptive
  Synchronization}.\hskip 1em plus 0.5em minus 0.4em\relax Red Hook, NY, USA:
  Curran Associates Inc., 2019.

\bibitem{monsoon}
Monsoon-solutions, ``High voltage power monitor.''
  \url{http://www.msoon.com/LabEquipment/PowerMonitor/}, December 2019.

\bibitem{websoc}
I.~Fette and A.~Melnikov, ``The websocket protocol,''
  \url{https://www.hjp.at/doc/rfc/rfc6455.html}, \text{RFC} 6455, IETF.
  Accessed April 4, 2021.

\bibitem{beutel2021flower}
D.~J. Beutel, T.~Topal, A.~Mathur, X.~Qiu, T.~Parcollet, P.~P.~B. de~Gusmao,
  and N.~D. Lane, ``Flower: A friendly federated learning framework,''
  \emph{arXiv preprint arXiv:2007.14390}, 2021.

\bibitem{pmlr-v54-mcmahan17a}
B.~McMahan, E.~Moore, D.~Ramage, S.~Hampson, and B.~A.~y. Arcas, ``{for
  federated learning in wireless networks},'' in \emph{Proceedings of the 20th
  International Conference on Artificial Intelligence and Statistics}, ser.
  Proceedings of Machine Learning Research, vol.~54.\hskip 1em plus 0.5em minus
  0.4em\relax PMLR, 20--22 Apr 2017, pp. 1273--1282.

\bibitem{hochreiter1997long}
S.~Hochreiter and J.~Schmidhuber, ``Long short-term memory,'' \emph{Neural
  computation}, vol.~9, no.~8, pp. 1735--1780, 1997.

\bibitem{deng2012mnist}
L.~Deng, ``The mnist database of handwritten digit images for machine learning
  research,'' \emph{IEEE Signal Processing Magazine}, vol.~29, no.~6, pp.
  141--142, 2012.

\bibitem{krizhevsky2009learning}
A.~Krizhevsky, G.~Hinton \emph{et~al.}, ``Learning multiple layers of features
  from tiny images,'' 2009.

\bibitem{Shakespeare}
S.~{Caldas}, S.~{Meher Karthik Duddu}, P.~{Wu}, T.~{Li},
  J.~{Kone{\v{c}}n{\'y}}, H.~B. {McMahan}, V.~{Smith}, and A.~{Talwalkar},
  ``{LEAF: A Benchmark for Federated Settings},'' \emph{arXiv e-prints}, Dec.
  2018.

\bibitem{Stisen2015SmartDA}
A.~Stisen, H.~Blunck, S.~Bhattacharya, T.~S. Prentow, M.~B. Kj{\ae}rgaard,
  A.~K. Dey, T.~Sonne, and M.~M. Jensen, ``Smart devices are different:
  Assessing and mitigatingmobile sensing heterogeneities for activity
  recognition,'' \emph{Proceedings of the 13th ACM Conference on Embedded
  Networked Sensor Systems}, 2015.

\bibitem{Yang20FedAVG}
Z.~Yang, M.~Chen, W.~Saad, C.~S. Hong, and M.~Shikh-Bahaei, ``Energy efficient
  federated learning over wireless communication networks,'' \emph{IEEE
  Transactions on Wireless Communications}, vol.~20, no.~3, pp. 1935--1949,
  2021.

\end{thebibliography}

\vfill

\end{document}